Tjaša Arčon, Marko Robnik-Šikonja, Polona Tratnik
# Large language models for folktale type automation based on motifs: Cinderella case study

## Abstract
Artificial intelligence approaches are being adapted to many research areas, including digital humanities. We built a methodology for large-scale analyses in folkloristics. Using machine learning and natural language processing, we automatically detected motifs in a large collection of *Cinderella* variants and analysed their similarities and differences with clustering and dimensionality reduction. The results show that large language models detect complex interactions in tales, enabling computational analysis of extensive text collections and facilitating cross-lingual comparisons.

## Zusammenfassung
Ansätze der Künstlichen Intelligenz werden in vielen Forschungsbereichen, einschließlich der Digital Humanities, adaptiert. Wir haben eine Methodologie für großangelegte Analysen in der Folkloristik entwickelt. Mithilfe von Machine Learning und Natural Language Processing erkannten wir automatisch Motive in einer umfangreichen Sammlung von Aschenputtel-Varianten und analysierten deren Ähnlichkeiten und Unterschiede durch Clustering und Dimensionsreduktion. Die Ergebnisse zeigen, dass Large Language Models komplexe Interaktionen in Erzählungen erfassen, wodurch die computergestützte Analyse umfangreicher Textsammlungen sowie die Durchführung sprachübergreifender Vergleiche ermöglicht wird.

## Résumé
Les approches de l'intelligence artificielle sont appliquées à de nombreux domaines de recherche, y compris les humanités numériques. Nous avons élaboré une méthodologie pour des analyses à grande échelle en folkloristique. Grâce à l'apprentissage automatique et au traitement du langage, nous avons détecté automatiquement des motifs dans un vaste corpus de variantes de Cendrillon et analysé leurs similitudes et différences par regroupement et réduction dimensionnelle. Les résultats montrent que les grands modèles de langage identifient des interactions complexes, facilitant l'analyse computationnelle et les comparaisons interlinguistiques.

## 1 Introduction
Large language models (LLMs) are transforming the way we obtain information, communicate, and work. At the same time, they are increasingly affecting many scientific areas, including those concerned with languages, humanities, and society. In the areas of computational linguistics, natural language processing (NLP), and natural language understanding (NLU), deep neural networks based on the transformer architecture have recently become dominant. Pre-trained LLMs such as BERT (Devlin et al. 2019), T5 (Raffel et al. 2020), GPT-3 (Brown et al. 2020), or its derivative ChatGPT, can solve many language tasks (e.g., information extraction, text classification, question answering, sentiment analysis, machine translation, topic detection, text simplification, summarization, etc.) better than traditional approaches (Wang et al. 2023) and are approaching human performance.

Fairy tales, folktales, and other ethnographic narrations, captivating narratives woven through time, are not merely timeless stories but reflections of the societies that produced and distributed them. The challenge is to automate the analysis of narrative structure with the examination of different versions of stories through time and cultures to allow for large scale analyses. It is characteristic of folktales that they appear in ever-new versions, with

**Tjaša Arčon**, University of Ljubljana, Faculty of Computer and Information Science, Slovenia.
E-mail: tjasa.arcon@fri.uni-lj.si
**Marko Robnik-Šikonja**, University of Ljubljana, Faculty of Computer and Information Science, Slovenia. E-mail: marko.robnik@fri.uni-lj.si
**Polona Tratnik**, University of Ljubljana, Faculty of Computer and Information Science; University of Ljubljana, Faculty of Arts, Slovenia; Institute IRRIS, Slovenia. E-mail: polona.tratnik@ff.uni-lj.si

individual tales sharing some common features, such as structure and motifs. Based on a set of common motifs or typical motifs and structures, the historical-geographic folkloristic school classified folktales into types. The aim of our research was to check whether we can automate the detection of motifs in folktales as defined by the ATU type index with LLMs. As an example, we took the tale of *Cinderella*, one of the most widespread and extensively studied folktales worldwide; however, it has never before been analysed with LLMs to try to automate the detection of motifs. With the LLM-based motif automation, we aim to: 1) analyse whether LLMs can help in determining the specifics of the tale variant, 2) recognize a certain folktale type on new examples of narratives, and 3) identify deviations and subversions of individual narratives concerning the normative model of the type. By showcasing the efficiency and accuracy of an LLM-based methodology on the *Cinderella* case, we open doors for its wider application in computational folkloristics.

To explore the potential of LLMs for identifying motifs in folktales automatically, we evaluated the zero-shot performance of GPT-4.5 Preview (OpenAI 2025) on a manually annotated sample of 13 *Cinderella* variants. We then applied the same methodology to a larger dataset of 77 folktales or fairy tales in English, originating from different geographical areas around the world, and then also to 33 *Cinderella* variants in Slovene, originating from the Slovene cultural area. The LLM was prompted to detect the presence or absence of motifs in each tale. We analyzed three distinct sets of motifs:

> 1. a set of 15 basic, specific motifs defined as typical for the ATU folktale type 510A;
> 2. a set of 18 motifs, composed of 15 basic motifs, 2 additional specific motifs, and 2 alternative motifs that replace one original basic motif in order to incorporate additional motifs of interest;
> 3. a set of 14 supermotifs as a generalized set of the basic motifs.

The LLM was prompted to detect motifs in each tale. The yes/no answers were transformed into matrices representing motif presence across tales. Using these matrices as input, we applied several clustering algorithms to find groups of similar tales in non-Slovene narratives. Subsequently, we examined how Slovene tales fit within the international corpus. Additionally, we employed text embeddings from the LabSE model and HDBSCAN clustering to assess semantic similarity across Cinderella variants.

The contributions of our work are as follows:
1. Development of a novel methodology, based on LLMs, for the automatic detection of folktale motifs and types.
2. Development of a text embedding-based methodology for finding groups of similar folktale variants, enabling their visualization and interpretation.
3. Evaluation of LLM-based folktale analysis on a large collection of *Cinderella* variants.
4. Testing both a basic and a generalized folktale type detector on a corpus of previously unanalysed and mostly unclassified Slovenian *Cinderella* variants in the Slovene language.

The paper is structured into six further sections. In Section 2, we present the background and related work. Section 3 outlines the selected *Cinderella* variants and the analysed motifs. In Section 4, we describe our methodology for the automatic detection of motifs and types using LLMs. Section 5 contains the results and evaluation of the proposed methodology applied to a collection of non-Slovene *Cinderella* folktales, followed by the mapping of Slovene variants onto this framework. In Section 6, we discuss the findings and address the limitations of our approach. Finally, in Section 7, we present the conclusions and implications for the wider application of the proposed methodology in folkloristics.

## 2 Background and related work

In this section, we first present the existing work on the classification of *Cinderella* variants and justify our selection of the dataset, followed by the existing works that utilize computational approaches to analyse folktales.

### 2.1 Cinderella variants in folkloristics

At the end of the nineteenth century, a British folklorist Marian Roalfe Cox collected three hundred and forty-five variants of *Cinderella*, *Catskin*, and *Cap'O Rushes* tales, and classified the versions of the folktale about Cinderella into five subtypes: A (*Cinderella: Ill-treated heroine, recognition by means of a shoe*), B (*Catskin: Unnatural father, heroine flight*), C (*Cap o' Rushes: outcast heroine*), D (*Indeterminate*) in E (*Hero-Tales*) (Roalfe Cox, 1893). In 1910, a Finnish folklorist of comparative folkloristics Antti Aarne built a comprehensive index of folktale types using the historical-geographical method, which was reviewed, revised, and expanded by the American folklorist Stith Thompson first in 1928 and then in 1961 (short AT index), and then further revised and expanded by the German folklorist Hans-Jörg Uther (short ATU index). In the type index of folktales, folktales about Cinderella are classified in group II., which comprises the so-called "*ordinary tales*," and within this group into category A., which comprises "*tales of magic*," to the section that includes tales with the supernatural helpers, namely under types 510 and 511. Type 510 includes folktales about *Cinderella* (510A) and *Cap O' Rushes* (510B). Subtype 510A refers to folktales with the following summarized description: "The two stepsisters. The stepdaughter at the grave of her mother, who helps her (milks the cow, shakes the apple tree, helps the old man). Three-fold visit to church (dance). Slipper test" (Thompson 1955–1958, 177). Type 510B comprises folktales "*The Dress of Gold, of Silver, and of Stars (Cap o' Rushes)*. Present of the father who wants to marry his own daughter. The maiden as servant of the prince, who throws various objects at her. The three-fold visit to the church and the forgotten shoe. Marriage" (Thompson 1955–1958, 177). Type 511 includes tales with "One- Eye, Two-Eyes, Three-Eyes stepdaughter who is abused by her mother" (Thompson 1955–1958, 178). Subtype 511A *The Little Red Ox* refers to folktales where an ox acts as a helper, while the helper can also be a cow. It also refers to a male hero, however, it can also comprise tales with a heroine (Thompson 1955–1958, 178).

A Swedish folklorist Anna Birgitta Rooth also dealt extensively with folktales about Cinderella, and classified the versions of this tale into the following subtypes: A I (*stepchildren, helpful cow, oriental spy motif, slaying the cow, burning of the bones, growing tree – provides food*), A II (*stepdaughter, task spinning, helpful cow, slaying of the cow/washing entrails, magic object, burying of the object, growing tree, fruit picking – marriage test, tree follows, wedding*). Both sets of folktales of type A correspond to the classification of ATU type 511. Type AB (*embroidered stepmother, helpful cow, task spinning, type Aarne 480, spy, grain sorting, fine clothing, visit to the feast, pursuit, lost shoe, show test, animal witness, wedding*). Type AB according to Rooth fits ATU types 511 and 510A. Type B (*stepdaughter, hearth abode, grain sorting, treasure tree, lost shoe, shoe test, and wedding*) corresponds ATU type 510A. Type BI (*death bed promise, deceased wife's ring marriage test, counter tasks, fine dresses, employment, wooden skirt (disguise), token objects named, visit to the feast, lovesick prince, recognition food, wedding*) corresponds ATU type 510B. And Type C (*stepson, helpful ox, ear cornucopia, food, flight at the slaughter, ride through three woods of metal, taboo-breaking of a twig, bull contest, killing of the ox, horn with herds of cattle, wedding, the suffering of the bread*) corresponds ATU type 511B (Rooth 1951, 14–22).

In our analysis, we primarily focused on the ATU type 510A, which is included by Rooth in the type B assembly, and by Roalfe Cox in the type A. Although the ATU Type Index remains today a key handy folkloristic tool for classifying and analysing folktales, critics have noted several shortcomings. Folktales are classified into types based on the set of specific motifs that each type contains. In support of the type index, Stith Thompson also produced a motif index (Thompson 1955–1958). Thompson himself admitted that "somewhat more than half of the types included consist of a single narrative motif" (Thompson 1964, 417). "The vast majority of animal tales (AT 1–299) are *both* single tale type numbers and single motif numbers" (Dundes 1997, 197). However, in this case, Thompson acknowledged, the problem of classification is relatively simple, while in many cases, types are "made up of a whole group of motifs and the question is continually presenting itself as to which of these motifs shall be used as the basis of classification" (Thompson 1964, 417). Alan Dundes established that the types are overlapping, which "occurs within *both* the conceptualization of motifs and tale types" (Dundes 1997, 197). In addition, Aarne based types partly on dramatis personae, which was for instance the basis for animal tales, where the principal actors were animal characters, however "[t]he reality of folktales, for example, demonstrates that the same tale can be told with either animal or human character," which is the reason to find the very same tale type listed twice in the AT index under two separate numbers (Dundes 1997, 197). Anna Birgitta Rooth acknowledged that individual motifs are often to be found interdependent upon other motifs in a given tale. Thus, she suggested using the notion of the "motif-complex" (Rooth 1951, 237-40). Bengt Holbek argued that the AT typology does not correspond to "the actual material collected in the field insofar as so-called types were often combined" (quoted in Dundes 1997, 196). Major criticism refers to the alleged Eurocentrism of the concepts of motif and tale type (Dundes 1997, 196). The ATU index has been criticized for structural and geographical bias (Jason 2006), since the selection and classification of types often reflect Western traditions while marginalizing other cultural variants. In addition, types are conceptualized around a supposedly central feature, such as "*supernatural helpers*," which might not even be the case in all of the variants of the type (Tratnik, 2023, 594). Roalfe-Cox's types of *Cinderella*, *Catskin*, and *Cap'O Rushes* are built upon Western variants, in particular, "type A. *Cinderella* upon Perrault's variant, type B. *Catskin* upon Basile's variant, and type C. *Cap o' Rushes* upon English variant of the tale". Similar holds true for the ATU subtypes ATU 510A, which is built upon the Grimm's variant and ATU 510B, which is built upon the English variant. Thus, the current types consolidate cultural hegemony in the case of ATU 510A (or Roalfe Cox A), especially by prioritizing Grimm's and Perrault's version of *Cinderella*, which adds another facet of criticism of the type index. In addition, it is also discussible if these variants are to be considered folktales, as it is known that the Brothers Grimm characteristically adapted these folktales (Zipes, 2010, 198), and the tales of the French court from the end of the seventeenth century are considered to be first cases of a literary genre of fairy tales (Zipes, 2001, xii). At the same time, recent scholarship has emphasized the multiplicity of *Cinderella* traditions across cultures and media (Hennard Dutheil de la Rochère et al. 2016), further questioning the privileging of a few Western versions as normative.

Through the analysis, we have established that the variants we collected rarely or almost never correspond to the precisely defined subtype ATU 510A, nor to types B according to Rooth or A according to Roalfe Cox, but they contain motifs from other types, or the motifs differ. We also established that folklorists initially cited an extensive set of motifs, probably recorded using the inductive method when examining the versions that are linked to a certain type, but later this set was reduced to a few typical motifs, which are supposed to define the type (see Uther, 2004). For example, Roalfe Cox associated the following motifs with type A: aid (various), animal witness, dead (or transformed) mother help, ear cornucopia, eating taboo, false bride, happy marriage, hearth abode, help at grave, helpful animal, heroine disguise and hero disguise, heroine flight and hero flight, ill-treated heroine and hero, lost shoe, lovesick prince, magic dresses, marriage tests, meeting-place, menial heroine and hero, mutilated feed, pitch trap, recognition by means of shoe or ring, revivified bones, shoe marriage test, laying of helpful animal, substituted bride, tasks, task-performing animal, threefold flight, token objects, villain nemesis.

The ATU index, edited by Uther, under type 510A lists the plot of the story with a few selected motifs, such as a cruel stepmother, helpful birds, a supernatural helper, a magic object received from mother, beautiful clothing from a supernatural being, a prince who falls in love with her, she has to leave the ball early, she loses one of her shoes, identification by the shoe, marriage.

> "A young woman is mistreated by her stepmother and stepsisters [S31, L55] and has to live in the ashes as a servant. When the sisters and the stepmother go to the ball (church), they give Cinderella an impossible task (e.g., sorting peas from ashes), which she accomplishes with the help of birds [B450]. She obtains beautiful clothing from a supernatural being [D 1050.1, N815] or a tree that grows on the grave of her deceased mother [D815.1, D842.1, E323.2] and goes unknown to the ball. A prince falls in love with her [N711.6, N711.4], but she has to leave the ball early [C761.3]. The same thing happens on the next evening, but on the third evening, she loses one of her shoes [R221, F823.2].
> The prince will marry only the woman whom the shoe fits [H36.1]. The stepsisters cut pieces off their feet to make them fit into the shoe [K1911.3.3.1.], but a bird calls attention to this deceit. Cinderella, who had been first hidden from the prince, tries on the shoe, and it fits her. The prince marries her" (Uther 2004, 293–294)."

This description of the plot corresponds to the Grimms' versions of *Cinderella* tales, in particular the one from the first edition of their Kinder- und Hausmärchen from 1812, where there is no motif of a father, who brings the gifts from the fairy to the girls, and where the motif of the tree is explicitly associated with the deceased mother (Grimm, 2014), wherein it seems that it was just this version taken as the normative for type 510A. In fact, only the segment "she obtains beautiful clothing from a supernatural being" is designed somewhat more generally regarding the tale of the Brothers Grimm, so that is captures also the help of the fairy, who otherwise appears in another most famous fairy tale about Cinderella written by Charles Perrault. The latter also generally corresponds to this scheme, although it is obvious that the plot type was not formed based on Perrault's version. In Perrault's version, Cinderella doesn't visit the grave of the deceased mother and does not receive the help of the birds, but turns for help to the godmother, who conjures up accessories for her participation at the ball. The version of Perrault also does not end with the mutilation of the feet, but the narrator says that the two sisters "did everything they could to get their feet into the slipper, but they could not do it" (Perrault 2009, 138). Uther then lists a wide range of variants from all over the world. Under the description of the plot, the index says: "This type is usually combined with episodes of one or more other types, esp. 327A, 403, 480, 510B, and also 408, 409, 431, 450, 511, 511A, 707, and 923" (Uther 2004, 294). The type is thus open to many other motifs in addition to those mentioned, as well as plot variants, although it is normatively determined in relation to Grimm's version of the fairy tale about Cinderella.

The AT Index provides a plot description for the entire Type 510 in the following most basic scheme: "*I. The persecuted heroine. II. Magic help. III. Meeting the prince. IV. Proof of identity. V. Marriage with prince. VI. Value of salt*". To these basic plot definitions, Aarne and Thompson added several implementation options. According to the given segments of the scheme, they list a total of seventy-four motifs for the common type 510 (Aarne and Thompson 1961, 175–176). Subtype 510A is summarized as follows: "*510A Cinderella. The two stepsisters. The stepdaughter at the grave of her own mother, who helps her (milks the cow, shakes the apple-tree, helps the old man*; cf. Type 480). *Three-fold visit to church (dance). Slipper test*" (Aarne and Thompson 1961, 177).

Compared to Uther's adaptation of the type 510A, the definition of the type 510 by Aarne and Thompson is a lot more general and thus comprises numerous versions. In addition, Uther compared to Aarne and Thompson chooses only some motifs as the typical ones for the type 501A, which testifies that the set of motifs listed by Aarne and Thompson for the type 510 is so comprehensive that it becomes too complex to give picture of a typical story and thus does not represent a set of the most typical motifs for the type, but rather numerous motifs that are connected with different versions. In such a manner this list of motifs is linked to numerous versions that are no longer connected with each other in the sense of forming subtypes. On the other hand, subtype 510A is defined too narrowly by Aarne and Thompson, as for instance Cinderella visits only the church, as well as it is specifically defined that it is the dead mother who helps the girl, while there are many various helpers in different versions, which might not be recognized as the reincarnations of the dead mother.

Libraries of the University of Missouri summarized type AT 510A or ATU 510A in the following assembly of motifs after Thompson: "*S31. Cruel stepmother. L55. Heroine's stepdaughter. B450. Helpful birds. D1050.1. Clothes produced by magic. N815. Fairy as helper. N711.6. Prince sees the heroine at the ball and is enamoured. C761.3. Taboo: staying too long at the ball. F823.2. Glass shoes. H36.1. Slipper test. K1911.3.3.1. False bride's mutilated feet*".

For our purpose of automating a tale type, we wanted to find a set of motifs that are typical of the type, which means that they typically appear in tales about Cinderella or which, in mutual combinations, form basic definitions of the type. On this basis, we decided to take the set of motifs as defined by ATU k10A as a starting point.

## 2.2 Computational methods in digital folkloristics

Several previous studies have employed computational methods to conduct research in folkloristics, but none has used large generative language models such as GPT-4 (OpenAI 2023) that exhibit significantly better performance in text analysis. An early folkloristic study integrating computational methods by Yarlott et al. (2016) set out to develop a system for motif extraction from narrative texts using a classical supervised machine learning approach and clustering techniques based on now outdated text representation with TF-IDF weighted Bag-of-Words (BoW). Thuillard et al. (2018) aimed to study and classify myths worldwide using phylogenetic and network analysis methods, as well as a clustering algorithm NeighborNet. A similar phylogenetic approach was adopted by Sakamoto Martini, Kendal and Tehrani (2023) who examined the global diversity and evolution of *Cinderella* tales and published the results of their phylomemetic case study of ATU 510/511. They tested Rooth's theories on a sample of 266 versions of *Cinderella* using Bayesian phylogenetic inference, phylogenetic networks (NeighborNet), and a model-based clustering method to uncover the historical relationships among types and subtypes, as well as the evolutionary processes that have shaped them (Sakamoto Martini, Kendal, and Tehrani, 2023). Abello et al. (2023) proposed network-based methods for disentangling complex interconnections among folklore motifs and tale types, highlighting both the potential and the pitfalls of computational classification. In contrast to the above-mentioned works, our study is not focused on diachronic analysis of *Cinderella* variations but presents a synchronic analysis of *Cinderella* tales type ATU 510A with the use of LLMs to identify the presence of motifs, followed by clustering based on the detected motifs.

Various studies have investigated deep learning methods in folkloristics. Pompeu et al. (2019) focused on classifying folktales into categories from the ATU classification system, employing a modified version of the Hierarchical Attention Network (HAN) deep learning model. Lô et al. (2020) explored how West African folk narratives could be generated with the help of recurrent neural networks (RNN) variant called Long Short-Term Memory (LSTM) networks, how they could be classified by comparing a classical machine learning classifier with BoW representation with an LSTM classifier, and how their narrative structure could be analyzed using classical machine learning classifiers like Naive Bayes, Support Vector Machines (SVM), and Logistic Regression. Similarly, Eklund et al. (2023) examined the classification of folktales into ATU tale types with the help of SVM. All these approaches use word-based text representations such as BoW and TF-IDF, which do not detect contextual differences and dependencies and, therefore, are not suitable for the analysis of complex questions. In our work, we use transformer-based LLMs, which, through extensive use of its self-attention mechanism, can detect long-term dependencies in the text with the text representation capturing contextual dependencies.

Recently, a handful of studies in folkloristics have attempted to use LLMs. Tangherlini and Chen (2024) utilized BERTopic to explore the interconnectedness of Hans Christian Andersen's fairy tales and travel writings. In addition, their methodology involved hierarchical density-based clustering (HDBSCAN) and dimensionality-reduction (UMAP) to group texts by topics. Meaney et al. (2024) aimed to classify Irish and Scottish Gaelic folktales into ATU categories and predict narrator gender employing several multilingual LLMs (mBERT, XLM-RoBERTa) and a monolingual Irish model (gaBERT). Hobson et al. (2024) used structured prompting with GPT-4 to extract story morals from a wide range of narrative genres, including folktales. These studies demonstrate the potential of LLMs to address traditional folkloristic questions with modern AI approaches.

Yarlott et al. (2024) used several LLMs (T5, FLAN-T5, GPT-2 and Llama 2) to classify motifs into four different types (motific, referential, eponymic, or unrelated) using a few-shot approach. The LLMs performed poorly as the best performance achieved 41% accuracy. The authors concluded that LLMs are not suitable for classifying motifs due to their lack of subtle cultural information. However, their research was carried out on news articles, which use motifs in a less structured way and not as consistently as fairytales, which are the primary focus in our research. Additionally, the LLMs used in the study were small compared to the state-of-the-art models such as GPT-4 and only snippets of articles with the motifs were provided to the LLMs due to architectural limitations of used LLMs. As human annotators, who did not possess expert knowledge, had difficulties in consistently labelling motifs, inconsistencies in annotation could influence the quality of the dataset and the evaluation results.

In our study, we explore whether particular motifs are present in different versions of *Cinderella*. The Thompson motif index was originally designed for folktales, which means that these motifs are well-documented in this domain. We use LLMs to identify motifs in tales, which are simpler and require less external knowledge for understanding than articles. Each tale is processed by LLMs in its entirety. Our research contributes to the advancement of digital folkloristics by developing an automated approach to large-scale analyses of motifs in tales, which enables further computational analyses.

## 3 Collected tales and detected motifs

In this section, we first outline the properties of the *Cinderella* dataset, then describe the analysed motifs.

### 3.1 *Cinderella* dataset

To prevent overfitting of the dataset with computational approaches and to assure reliable and generalizable results, initial evaluations of LLM responses were performed on 13 *Cinderella* variants in English by comparing the responses with human evaluations of the motifs present in the tales, before expanding the analysis to the full dataset.

As the initial results were satisfactory, we subsequently used the LLM to find motifs in 77 Cinderella variants from diverse geographical regions worldwide, all translated to English. We compiled the dataset by first selecting 17 tales representing ATU 510A listed in the digital ATU index of the Libraries of University of Missouri, which comprise the most known tales, such as those written by the brothers Grimm and Charles Perrault, as well as originating from different parts of the world. We extended our collection with tales referred to as typical for the ATU 510A type from various sources, such as those listed by D. L Ashliman, University of Pittsburgh. The tales from our collection are written in English and originate from five continents: 44 from Europe, 18 from Asia, 11 from the Americas, three from Africa and one from Antarctica.

Finally, we applied the procedure to 33 *Cinderella* variants from the Slovene-speaking regions, written in Slovene, most of which have not yet been classified as ATU 510A, but which we determined, upon review, belonged to this type. We collected all the Slovene variants that we identified as belonging to the *Cinderella* tradition after reading them.

## 3.2 Motif selection and generalizations

To analyse the motif inventory in *Cinderella* tales, we performed our analysis in three stages, each focusing on a different set of motifs. We first analysed a set of 15 motifs, as specified by Uther's narrow definition of the folktale type ATU510, identified by Thompson. After reviewing the responses provided by GPT-4.5-Preview (OpenAI 2025), we concluded that several narrowly defined motifs, which were not part of the original inventory, could provide valuable insights into the analysis of the Cinderella tales. As a result, we decided to add some changes to the set of narrowly defined motifs (see Table 1 for more details). We supplemented the motif of a cruel stepmother (Thompson's *S31.*) by incorporating the motif of incestuous first parents (Thompson's *A1273.1.*) and expanded the motif of a stepdaughter heroine by adding the complementary motif of *a stepson hero*, a motif that we adapted from Thompson's motif "*L10. Victorious youngest son*". In this way, we added some motifs that, according to the previous classifications, would otherwise belong to other types or subtypes of the *Cinderella* tale, such as the motif of the incestuous father. These two motifs differentiate those tales that speak about incestuous father and those that speak of a male main character from other tales that exclude these motifs, which was also the basis for special categorization of these tales for some researchers (Roalfe Cox 1893; Rooth 1951). The motif of the incestuous father is present in some versions of the classical *Cinderella* type, i.e., in ATU subtype 510A (type A according to Roalfe Cox), although according to the ATU index it appears in subtype 510B, while in Roalfe Cox's classification it appears in type B. We also modified the motif of helpful birds, which refers narrowly to birds, while many variants feature other animals in the same role as helpers, especially a cow (which is also the helper in AT 510A) or an ox. Because this element also speaks of significant differences between the tales, we considered it important to implement the differentiation between the wild helping animals (such as the birds, the motif Thompson's *B450*) and domestic helping animals (such as a cow or an ox) to better reflect the narratives' distinctions associated with different animals across various versions. The changes we implemented in the ATU 510A set of motifs were thus meant to get a better representation of the similarities and differences between the variants examined. The adapted set consisting of 18 specific motifs served as the second set of motifs used in our analysis.

After examining motif frequencies within our *Cinderella* dataset, we found that the motif inventory did not adequately represent the majority of the tales. Only three of the original 15 motifs appeared in more than two thirds of the *Cinderella* tales, while most motifs occurred in fewer than one third of the stories (see Figure 1 for detailed frequencies of individual motifs). Across all *Cinderella* versions, the most frequently recurring motif is that of a cruel stepmother, appearing in 75% of the tales. The motif of a stepdaughter heroine is present in 70% of the *Cinderella* variants and the motif of a shoe/slipper test appears in 66% of the narrations. The remaining motifs occur less frequently. An example of a motif that was too narrowly defined is the motif of glass shoes, which is present in only a few variants, for example, in the fairy tale written by Charles Perrault, for which Honoré de Balzac objected that the shoes must have been made of fur (because of the similarity between the French words *verre* (glass) and *vair* (antique for squirrel fur) (Betts, 2009, 202–203).

Because the original 15 motifs, typically expected to compose the Cinderella tale, were too specific to be considered normative for the type (such as for instance Thompson's *F823.2.* glass shoes), we searched for a more normative set of motifs that would cover a wider range of *Cinderella* variants. We identified more general motifs directly superordinated to the initial ones, following Thompson's classification. We created an alternative set consisting of 14 generalized motifs, or supermotifs motifs (see Table 1 for more details). In this process, most of the original motifs were generalized, and two narrowly defined motifs (*N711.6. "Prince sees heroine at ball and is enamored"* and *N711.4. "Prince sees maiden at church and is enamored"*) from the original ATU 510A set were merged into a single motif (*N710. "Accidental meeting of hero and heroine"*) to better represent their shared narrative function. The broadly defined set of general motifs was based on the additional comments provided in the LLMs' responses. LLMs generated these comments whenever the motifs in the tales deviated from the original set of narrowly defined motifs. In this way, we were able to identify recurring patterns in tales. The analysis of the supermotifs' frequencies revealed improved coverage, with five supermotifs occurring in more than two thirds of the tales and only three appearing in fewer than one third of the stories (see Figure 1 for detailed frequencies of individual supermotifs).

Using computer processing, we analysed both specific and broad motifs, extracting the basic general motifs of the *Cinderella* tale type that appear in most *Cinderella* variants, while also examining the distinctive features of individual variants using detailed motif definitions, which provided insight into connections between the variants.

**Table 1.** Three sets of motifs in *Cinderella* tales.

| ATU 510A Uther's motifs, according to Thompson | Additional ATU 510A motifs | Supermotifs, according to Thompson |
|---|---|---|
| *S31.* **Cruel stepmother** | *S31.* **Cruel stepmother** | †S0--S99. **Cruel relatives.** |
| | *A1273.1.* **Incestuous first parents** | |
| *L55.* **Stepdaughter heroine** | **Stepdaughter heroine** | †L0--†L99. **Victorious youngest child.** |
| | **Stepson hero** | |
| *B450.* **Helpful birds** | **Helpful domestic animals** | †B400--†B499. **Kinds of helpful animals.** |
| | **Helpful wild animals** | |
| *D1050.1* **Clothes produced by magic** | | *D1050.* **Magic clothes.** |
| *N815.* **Fairy as helper** | | *N810.* **Supernatural helpers.** |
| *D815.1.* **Magic object received from mother** | | *D810.* **Magic object a gift.** |
| *D842.1.* **Magic object found on mother's grave** | | *D840.* **Magic object found**. |
| *E323.2.* **Dead mother returns to aid persecuted children** | | *E320.* **Dead relative's friendly return.** |
| *N711.6.* **Prince sees heroine at ball and is enamored** | | *N710.* **Accidental meeting of hero and heroine**. |
| *N711.4.* **Prince sees maiden at church and is enamored** | | |
| *C761.3.* **Tabu: staying too long at ball. Must leave before certain hour** | | *C750.* **Time tabus.** |
| *R221.* **Heroine's three-fold flight from ball** | | *R220.* **Flights (as hurried escapes)**. |
| *F823.2.* **Glass shoes** | | *F820.* **Extraordinary clothing and ornaments.** |
| *H36.1.* **Slipper test. Identification by fitting of slipper** | | †H0--†H199. **Identity tests: Recognition.** |
| *K1911.3.3.1.* **False bride's mutilated feet.** In order to wear the shoes with which the husband is testing the identity of his bride, the false bride cuts her feet. She is detected | | *K1910.* **Marital impostors.** |

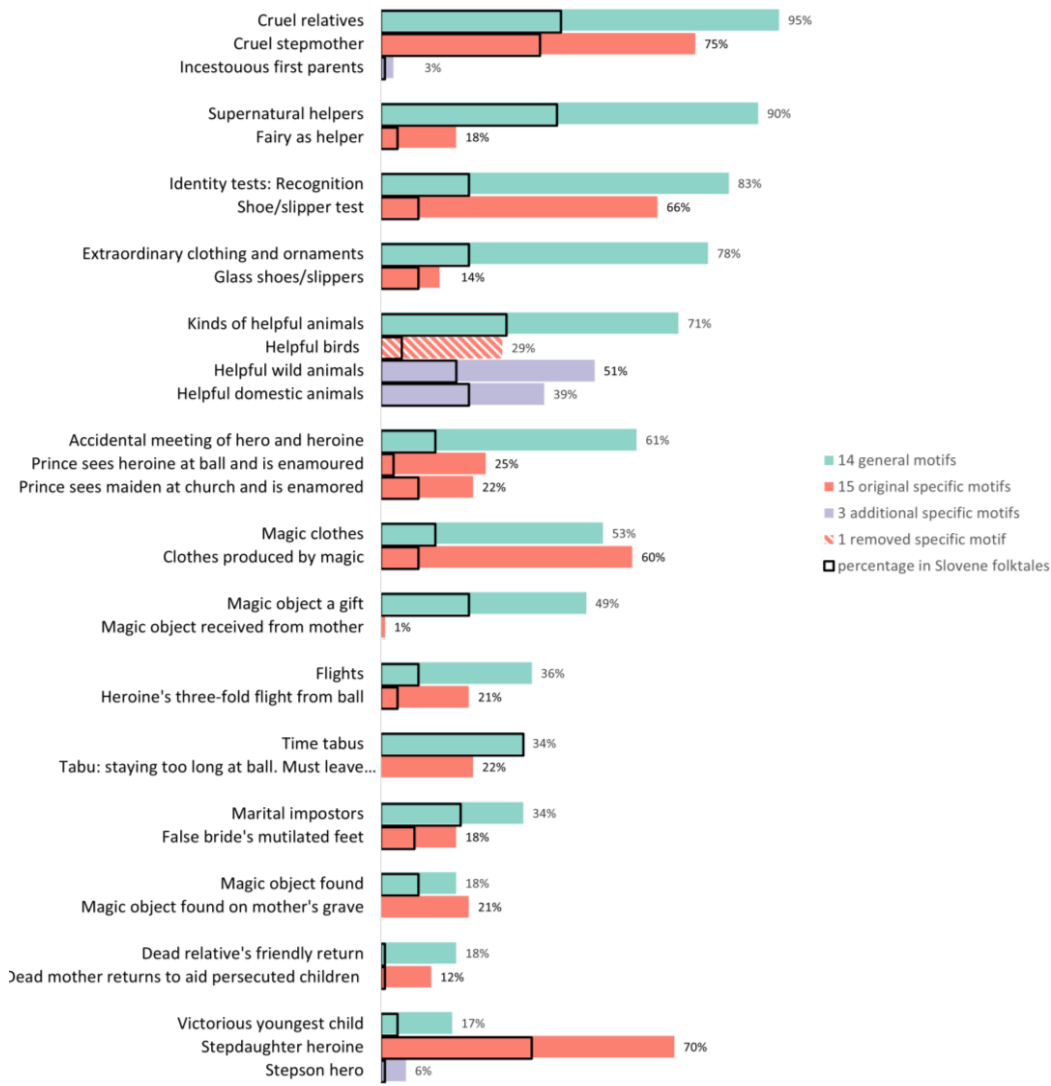

**Figure 1.** Distribution of motifs in foreign and Slovene folktales.

## 4 Methodology for motif and type detection

In this section, we present the computational methodology used in our analysis. In Section 4.1, we describe our approach to detecting motifs in individual tales. In Section 4.2, we explain how we identified groups of non-Slovene tales with similar motifs using clustering algorithms. In Section 4.3, we extend the analysis to Slovene Cinderella variants by mapping them onto non-Slovene clusters. Finally, in Section 4.4, we briefly discuss an alternative method for finding similar tales by clustering text embeddings derived from the multilingual LaBSE model. We report the results of these methods in Section 5.

## 4.1 Motif detection with LLMs

Large language models (LLMs) are artificial intelligence systems that process and generate text. The core method of interacting with them is through prompts. Several tasks can be achieved through prompting such as text classification, assigning tags to individual words or phrases in a text, or text generation (Kamath 2024). In our study, we automated identification of motifs in tales with the help of LLMs. To explore the potential of using LLMs for identifying motifs and to test their alignment with human judgements, we first made an initial evaluation of state-of-the-art LLM, namely GPT-4.5-Preview (OpenAI 2025) performance. We tested how accurately GPT-4.5-Preview detected motifs on a sample of 13 tales, where the presence of ATU 510A motifs, typical of *Cinderella*, had been manually annotated. We compared GPT-4.5-Preview responses with manually annotated motifs. Since the initial results were satisfactory, we subsequently applied the automatic procedure to a larger dataset consisting of 110 tales, 77 in English and 33 in Slovene. The two groups were submitted to the LLM and processed separately.

To determine the presence of motifs in each tale, we employed a zero-shot approach using the GPT-4.5-Preview model via API access, meaning that the model received only a prompt without additional examples. We instructed the model to identify all listed motifs within the given tale. For this purpose, system and user messages were included in the zero-shot prompt, along with the tale text. This process was repeated for all versions of *Cinderella*. We provided the same prompt in English for tales written in both English and Slovene.

The presence of motifs for both general and narrowly defined motifs was tested using the same prompt, with only the listed motifs modified accordingly. The following prompt was used to find general motifs:
- system message: "You are a helpful assistant, skilled in finding motifs in tales. Follow the instructions carefully."
- user message: " Find motifs in the following story. For each motif, return 'yes' if the motif is present in the story, return 'no' if the motif is absent. If the motif is modified, add a very short comment next to yes or no. Do not write the motif. Choose among the following motifs: 1. Cruel relatives, 2. Victorious youngest child, 3. Kinds of helpful animals, 4. Magic clothes, 5. Supernatural helper, 6. Magic object a gift, 7. Magic object found 8. Dead relative's friendly return, 9. Accidental meeting of hero and heroine, 10. Time tabu, 11. Hurried escape 12. Extraordinary clothing and ornaments, 13. Identity test: Recognition, 14. Marital impostor. Example: 1. yes 2. no etc.

The yes/no responses obtained from the LLM model were converted into a binary $n \times m$ matrix, where $n$ is the number of fairy tales and $m$ the number of motifs. In the matrix, the value of 0 indicated the absence of a motif and the value of 1 indicated its presence. When the response included an additional short comment, only the yes/no answer was translated into 0 or 1, the comment was omitted and analysed separately. As we performed three analyses based on different sets of motifs, six matrices were created: three for the non-Slovene tales and three for the Slovene tales. The first pair of matrices consisted of the original 15 narrowly defined ATU 510A motifs, the second pair included 18 narrowly defined motifs, and the third pair was comprised of 14 more general motifs (see Table 1 for details). The resulting three binary matrices for 77 non-Slovene tales were used as input for several clustering algorithms that automatically grouped Cinderella tales based on motif similarity. Subsequently, the three matrices for the Slovene tales were used to determine how these narratives fit within the clustering results based on the non-Slovene tales.

The short comments in LLM responses guided the evaluation of the initial set of narrowly defined motifs. As the LLM consistently included comments in the responses whenever modifications of original motifs were observed, we used these to identify which motifs were actually applied in the tales. These comments helped refine the initial set of specific motifs and provide a foundation for developing the broadly defined set of general motifs.
We report the results of the described methodology in Section 5.1.

## 4.2 Identifying type patterns in the motif space using clustering
Clustering is the process of grouping similar elements together so that the elements within the same cluster are similar, while elements in different clusters are distinct (Han et al. 2012). We performed clustering on narratives based on the presence or absence of motifs in each tale to determine which groups of tales share similar motif patterns. We implemented clustering on 77 non-Slovene tales, using motifs as features, with three different sets of motifs, described in Sections 3.2 and 4.1. The first set consisted of the original 15 motifs of the 510A *Cinderella* ATU type. The second set contained 18 motifs that were a combination of the original narrowly defined motifs and their extensions. The final set included 14 more general motifs.

Through this approach, we tried to group the tales based on their motif similarity and, consequently, identify underlying patterns and relationships among the *Cinderella* variants. To identify similarity patterns between different tale variants, we applied several clustering algorithms, including K-means (MacQueen 1967), KMedoids (Kaufmann and Rousseeuw 1990), Agglomerative clustering (Ward 1963), and DBSCAN (Ester et al. 1996) clustering. To select the best clustering, we computed the Silhouette coefficient (Rousseeuw 1987), which evaluates the quality of the clustering results based on how well data points within a cluster are separated from those in other clusters.

As the values of Silhouette coefficient were quite low, we repeated the clustering procedure on data with less dimensions. Since clustering algorithms are sensitive to high-dimensional data, we used a standard approach to reduce this sensitivity. Before applying clustering for the second time, we used the UMAP (Uniform Manifold Approximation and Projection) method to reduce data dimensionality while preserving similarity from the original high-dimensional space as much as possible. We repeated all four clustering procedures on the reduced dimensionality space. To select the best clustering, we again computed the Silhouette coefficient and concluded that, according to this criterion, K-means clustering produced the best results. This algorithm was then applied to all three motif subsets. We report the results of the described methodology in Section 5.2.

### 4.3 Mapping Slovene tales onto defined clusters

After performing the clustering of non-Slovene tales for each of the motif sets, we added 33 Slovene *Cinderella* versions to the analysis to identify which existing clusters of tales express strongest similarity to each of the Slovene narratives. The procedure was repeated for each of the motif sets, first for the original 15 motifs, then for the extended 18 specific motifs, and finally for the more general set of 14 motifs. This approach allowed us to classify the *Cinderella* variants in Slovene, the majority of which had not been classified previously, and to perform a further analysis of such classification. We examined whether the Slovene tales were similar to the non-Slovene ones and how closely they fit the identified specific or general patterns. We report the results of this procedure in Section 5.3.

### 4.4 Applying text embeddings for similarity analysis

In addition to analysing tale patterns and relationships through clustering based on motif-similarity, we explored semantic similarity among tales using clustering on text embeddings. To achieve this, we employed the LaBSE (Language-agnostic BERT Sentence Embedding) model (Feng et al. 2020), which encodes text into high-dimensional vectors, generating embeddings that represent the semantic meaning of the text. LaBSE is a multilingual model that is suited for cross-lingual tasks, supporting more than 100 languages, including English and Slovene. Prior to clustering, the generated embeddings were reduced in dimensionality using UMAP. The resulting lower-dimensional embeddings were then clustered with HDBSCAN (Hierarchical Density-Based Spatial Clustering of Applications with Noise) clustering algorithm (McInnes et al. 2017), which is suitable for high dimensional spaces. We report the results of the described methodology in Section 5.4.

### 5 Evaluation and Results

In this section, we present the results of the developed methodological approaches from Section 4. In Section 5.1, we describe the results for detection of motifs in individual tales. In Section 5.2, we outline identified groups of non-Slovene tales based on similarity of motifs and using clustering algorithms. In Section 5.3, we describe how Slovene *Cinderella* variants map to previously described non-Slovene clusters. In Section 5.4, we describe clusters obtained from an alternative text representation method obtained from the multilingual

LaBSE model. In Section 5.5, we describe the geographical distribution of the obtained clusters.

## 5.1 Success of motif detection with LLM

During the testing stage, we assessed the performance of the GPT-4.5-Preview large language model for motif detection using an initial dataset of 13 tales. Model performance was evaluated using accuracy, which measures the proportion of correct answers (true positives and true negatives) compared to the total number of manually annotated examples. The initial evaluation of 13 different fairytales was successful, with 98% of the motifs correctly identified in comparison to manually annotated motifs. We tested GPT-4.5-Preview with 15 original ATU 510A motifs. This high accuracy score gave us confidence to apply the model to all the tale variants and analyse the obtained results. It is worth noting that human evaluation of even the initial 13 tales took considerable time, and a large-scale evaluation of all the 110 variants would be tedious.

As previously mentioned, we extended the original set of narrowly defined motifs and created an additional set of broader motifs, drawing on the frequency of each of the 15 Thompson ATU motifs for *Cinderella* across 77 non-Slovene variants, as well as on the generated LLM yes/no responses for the initial 13 non-Slovene variants. In most cases where the motifs in the tale were modified, GPT-4.5-Preview added a short comment to the yes/ no response, specifying the type of motif. For example, in the case of the "glass shoe" motif, the negative answers contained a short description of the shoe used in the fairytale. Or another example, when asking about the "birds as helpers" motif, the negative LLM response included a clarification that another animal was present such as, for instance, "a cow" or "a bull", which proves that LLMs can successfully differentiate even between closely related motifs and in this way contribute to a more nuanced motif analysis.

## 5.2 Identified type patterns through clustering

When detecting fairytale similarity patterns, we first selected the clustering algorithm, using the Silhouette Coefficient (Rousseeuw 1987). This metric measures how similar an object is to others in its own cluster compared to those in other clusters, assessing both the cohesion within clusters and the separation between them. The Silhouette Coefficient ranges from -1 to 1, with higher values indicating more coherent clusters. Values close to 1 indicate that the elements in a particular cluster are very similar to their own cluster (i.e. central in their own cluster), while at the same time dissimilar to elements in other clusters, whereas values close to 0 indicate that elements in a specific cluster are not typical of their own cluster (i.e. mostly lie near cluster boundaries). Additionally, we used the Silhouette Coefficient to determine the optimal number of clusters, representing groups of similar tales. By testing multiple clustering configurations with several possible numbers of clusters, we selected the one with the highest Silhouette Coefficient.

Among the clustering algorithms tested, K-means clustering produced the highest Silhouette Coefficient. However, Silhouette Coefficients were low, which indicated poor cluster separation, so we repeated the clustering process on the same data, but first reduced its dimensionality using UMAP. We applied this dimensionality reduction technique, expecting it to remove redundant features, in our case, uninformative motifs, to improve cluster separation and lead to better-defined groups of tales.

The use of UMAP for dimensionality reduction significantly improved the clustering results, as shown by high values of Silhouette Coefficients. The best clustering results after UMAP dimensionality reduction were again achieved using the K-means algorithm. K-means clustering with the original set of 15 motifs produced two well-defined final clusters with a high Silhouette Coefficient (s=0.9), showing a high degree of separation between clusters. When clustering the narrowly defined motif set with additional motifs, the number of clusters remained at two, with the Silhouette Coefficient staying high (s=0.81). Performing K-means clustering on the broadly defined 14 motifs split the tales into four clusters with a moderately good Silhouette coefficient (s=0.6), suggesting that although tales within each cluster share clear similarities in motif structure, the separation from other clusters is moderate.

Examining our clustering results, and the motifs contained in each cluster, we identified the main motif patterns of *Cinderella*. We compared groupings derived from specific motifs with those based on broader, superordinate motifs. Clustering based on the original, narrowly defined set of 15 motifs divided the tales into two clusters. These were identical to those obtained using the narrowly defined set of 18 motifs, which included the modified motifs of particular interest, (1) the motif of an incestuous father and a stepson hero, and (2) the substituted motif of helpful birds with the motifs of helpful domestic and wild animals.

Clustering both sets of narrowly defined motifs divided the tales into one larger cluster with 58 tales and one smaller cluster with 19 tales as illustrated in Figure 2. The most common motifs in the larger cluster, appearing in at least two thirds of the tales in this cluster are: a cruel stepmother, a stepdaughter heroine, a shoe/slipper test and clothes produced by magic. In contrast, no motifs are present with high frequency in the smaller cluster. The only motifs woven into the narrative in at least one third of the tales in the smaller cluster are: shoe/slipper test and clothes produced by magic. In the extended set of 18 motifs, illustrated in Figure 3, helpful wild animals and helpful domestic animals appear each in around half of the tales.

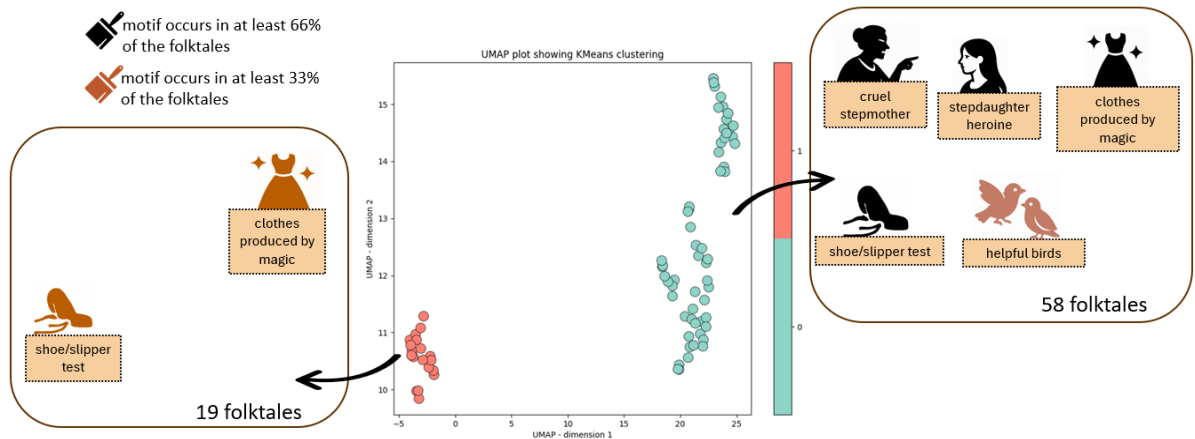

**Figure 2.** K-Means clustering results for 15 original narrowly defined motifs

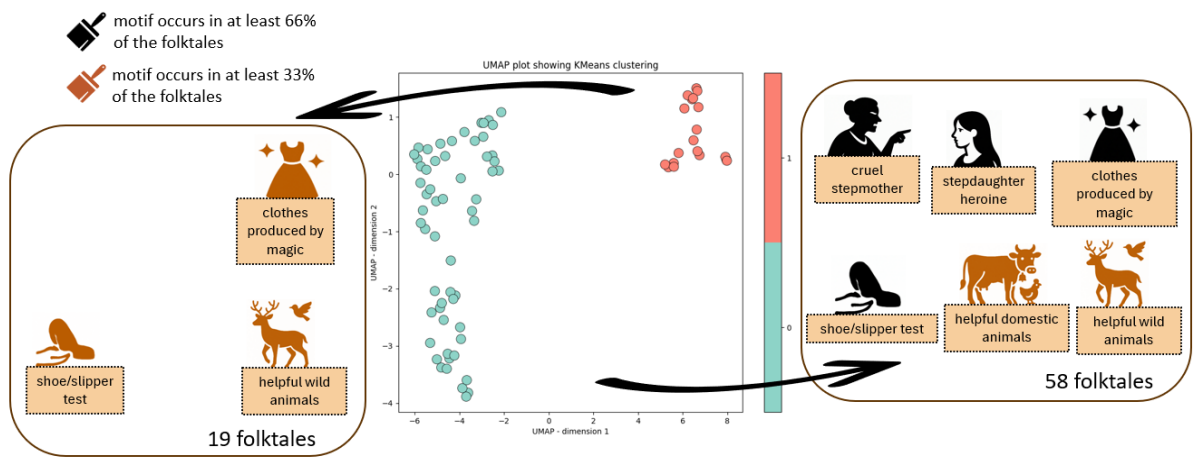

**Figure 3.** K-Means clustering results for 18 narrowly defined motifs

When performing K-means clustering using superordinate motifs, the tales were divided into four groups as illustrated in Figure 4. Two highly frequent motifs occurred in all clusters:

cruel relatives and supernatural helpers. Two clusters, one comprising 33 tales and the other 13, shared eight general motifs: cruel relatives, supernatural helpers, helpful animals, magic clothes, accidental meeting of hero and heroine, extraordinary clothing and ornaments, identity test and an object given as a gift. They differed mainly in motif frequency and in the presence of the motifs of marital impostors and magic object found in the larger cluster, and the motifs of flight and time taboo in the smaller cluster. The second-largest cluster, with 23 tales, also overlapped significantly with the previously mentioned two clusters, but was distinguished by the absence of magic clothes. The smallest cluster, containing only five narratives, was unique because, in addition to the four motifs common to all clusters, it also featured the motif of a victorious youngest child.

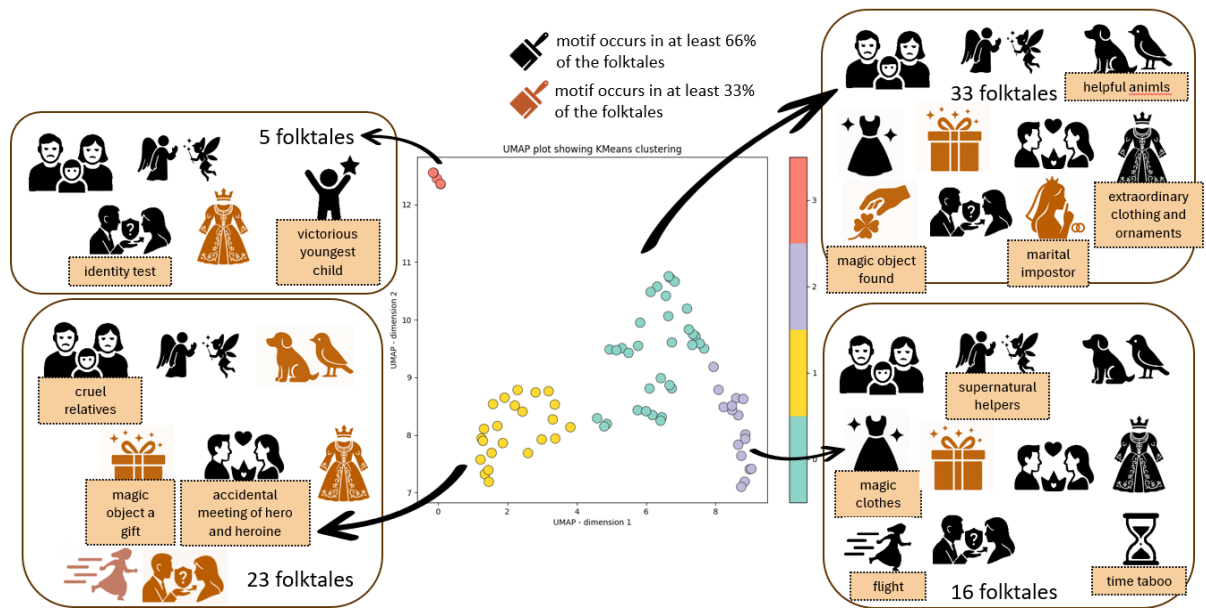

**Figure 4.** K-Means clustering results for 14 general motifs

## 5.3 Identified type patterns in Slovene

Mapping Slovene tales onto the established type patters derived from non-Slovene *Cinderella* versions revealed that Slovene versions of *Cinderella* correspond to a high degree to the specific and general motif types found by clustering of non-Slovene tales.

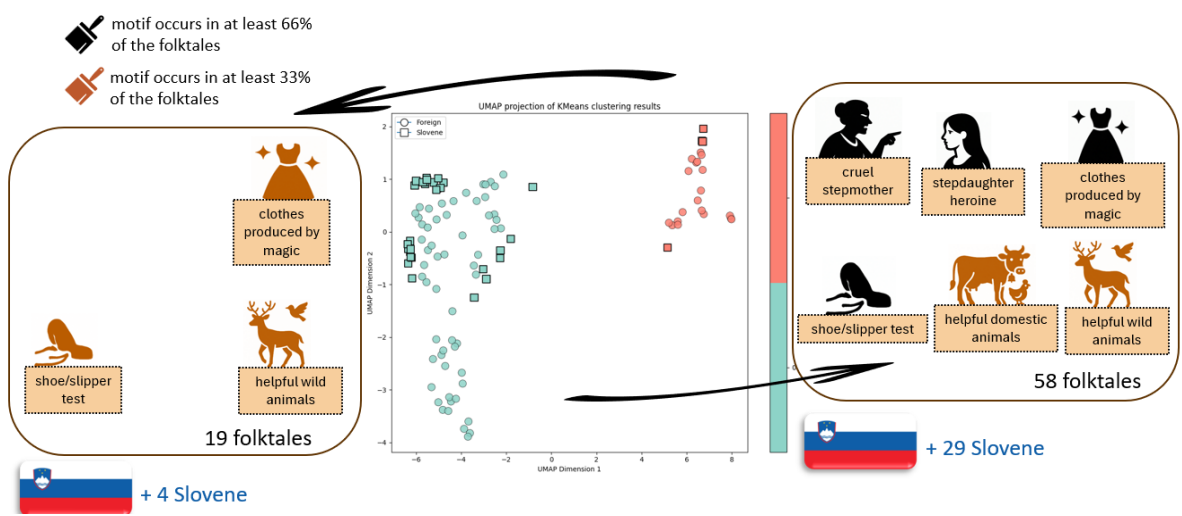

**Figure 5.** *Cinderella* variants in Slovene – clustering results for 15 narrowly defined motifs

Clustering results based on the presence and absence of 15 original motifs and the extended set of 18 specific motifs gave similar results (see Figures 5 and 6). Most Slovene tales (88%) fall into the largest cluster characterized by the motifs of a cruel stepmother, a stepdaughter heroine, a shoe/slipper test and clothes produced by magic. Another motif that appears in at least one third of the tales is the motif of helpful birds in the first set and the motifs of helpful wild animals and helpful domestic animals in the second set. Only four *Cinderella* variants in Slovene correspond to the cluster with no particularly distinctive motifs.

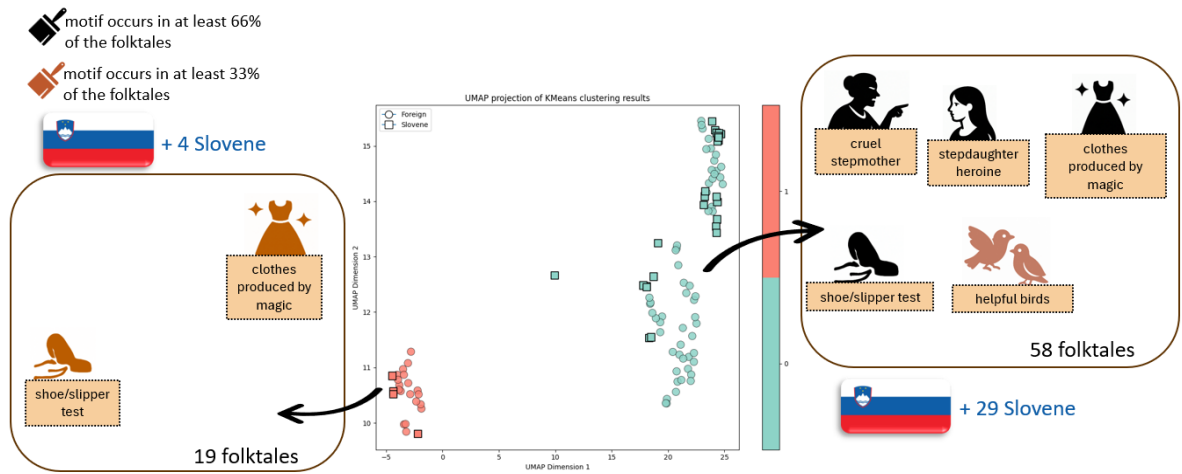

**Figure 6.** *Cinderella* variants in Slovene – clustering results for 18 narrowly defined motifs

In the clustering results based on general motifs, illustrated in Figure 7, one cluster, sets itself apart, with 20 Slovene tales assigned to it. It is interesting that this is the second largest cluster with non-Slovene tales, and not the largest. The most common motifs in this cluster are cruel relatives, supernatural helpers, and an accidental meeting of the hero and heroine. Helpful animals, magic object as a gift, flight, extraordinary clothes and ornaments, and an identity test are also frequent. What stands out is the absence of the motifs of magic clothes, time tabus and marital impostor. The rest of the *Cinderella* variants in Slovene are incorporated into the clusters where magic clothes appear at a high frequency together with the majority of other general motifs, which distinguishes the two clusters from other clusters. Nine tales form part of the largest cluster where the motifs of marital impostor and magic object found also occur, while four tales can be found in the cluster that exhibits a high frequency of tales with the motifs of flight and time taboo. No Slovene variants are incorporated into the cluster with a victorious youngest child.

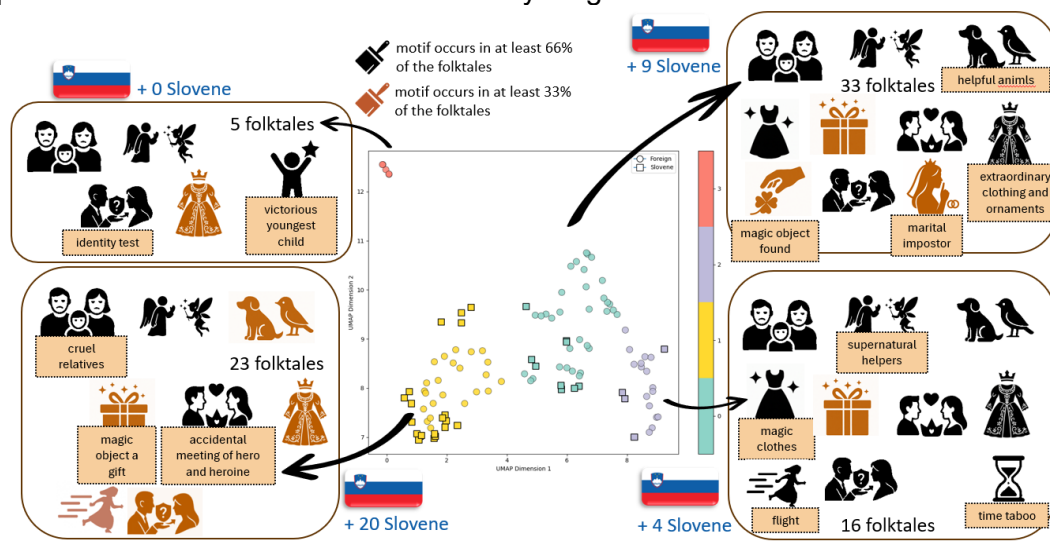

**Figure 7.** *Cinderella* variants in Slovene – clustering results for 14 general motifs

## 5.4 Patterns in the text embeddings space

We conducted an additional semantic analysis of the tales using LaBSE embeddings in combination with HDBSCAN clustering to see if further patterns and relationships between different versions of the Cinderella tales could be discovered based on textual similarity. To begin, we applied UMAP to reduce the dimensionality of the data. The best clustering configuration was then selected on the basis of the highest Silhouette coefficient. HDBSCAN clustering, performed on the embeddings, divided the non-Slovene tales into 26 clusters. However, we considered such a high number of clusters too fragmented to allow for a meaningful categorization of *Cinderella* variants. In addition, the clustering results obtained through similarity search on LaBSE embeddings differed substantially from the clustering results based on motif similarity. Consequently, no further analysis was conducted on the tales belonging to each of the three groups.

## 5.3    Geographical distribution

Figure 8 presents the geographical distribution of the non-Slovene tale clusters grounded in narrowly defined motifs. As the clustering results of the two sets based on specific motifs are identical, we present them only once. Both clusters represent a mixture of tales from different continents. The largest cluster consists predominantly of European and Asian tales with a smaller contribution from American tales and one tale from Antarctica. In contrast, the smaller cluster does not contain primarily European and Asian tales, but instead represents a blend of variants from America, Africa, and Europe. Notably, the smaller cluster contains only one Asian tale. Although European and American tales dominate in both clusters, one is further characterized by a presence of Asian narratives, whereas the other features African narratives. Tales in the former cluster typically include the motifs of a cruel stepmother, a stepdaughter heroine, a shoe/slipper test and clothes produced by magic. In contrast, tales in the latter cluster do not display any typical common motifs, apart from the motifs of shoe/slipper test and clothes produced by magic, which appear in roughly half of its tales. When the second set of 18 specific motifs is applied, the motif of wild animals also emerges in nearly half of the tales from the latter cluster.

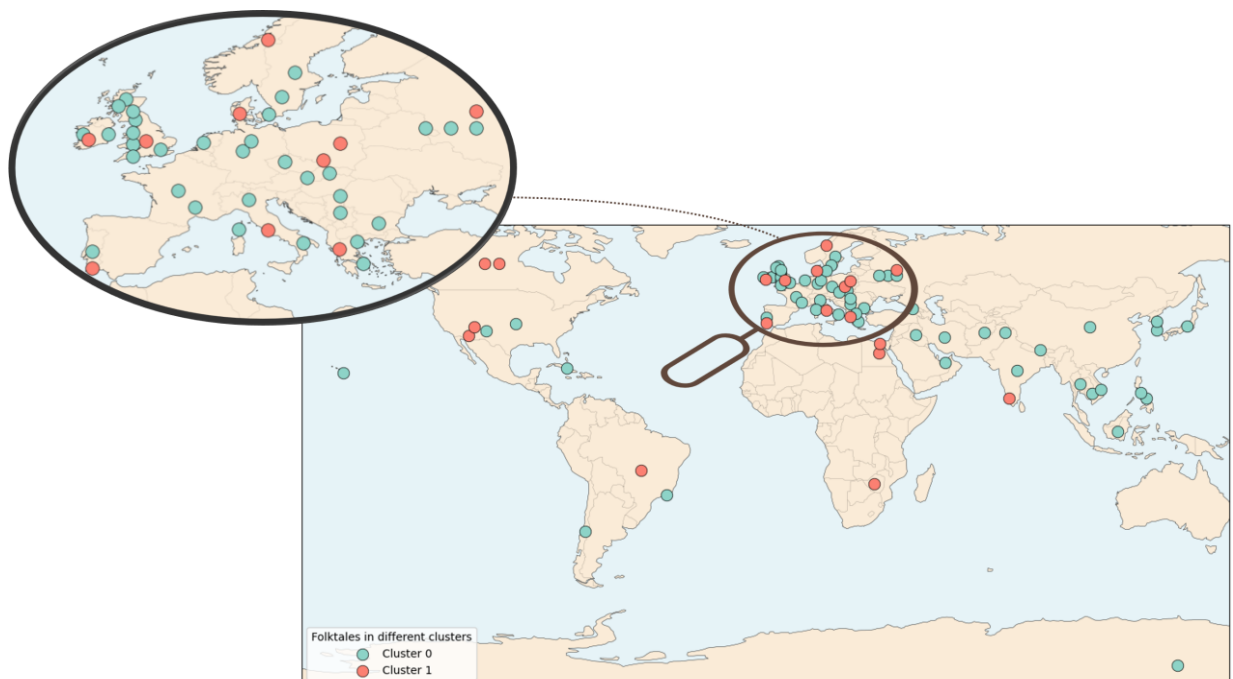

**Figure 8.** The geographical distribution of clusters based on narrow motifs

When we use the set of general motifs to group non-Slovene tales, each cluster likewise consists of tales originating from different continents. Figure 9 presents the geographical distribution of the non-Slovene tale clusters grounded in broadly defined motifs. The two largest clusters are composed predominantly of European and Asian tales, with the largest also incorporating a few American tales. The main difference between them is that the motif of magic clothes does not appear in the cluster with fewer tales. The third cluster includes mostly European and American variants. It shares most of the common motifs with the largest cluster, but its highly distinctive feature is the motif of time taboo, present in nearly all the tales in the cluster, yet comparatively rare in the other clusters. The smallest cluster based on the set of general motifs incorporates only five narratives, three American, one European and one African. Its most distinctive feature is the presence of the motif of a victorious youngest child.

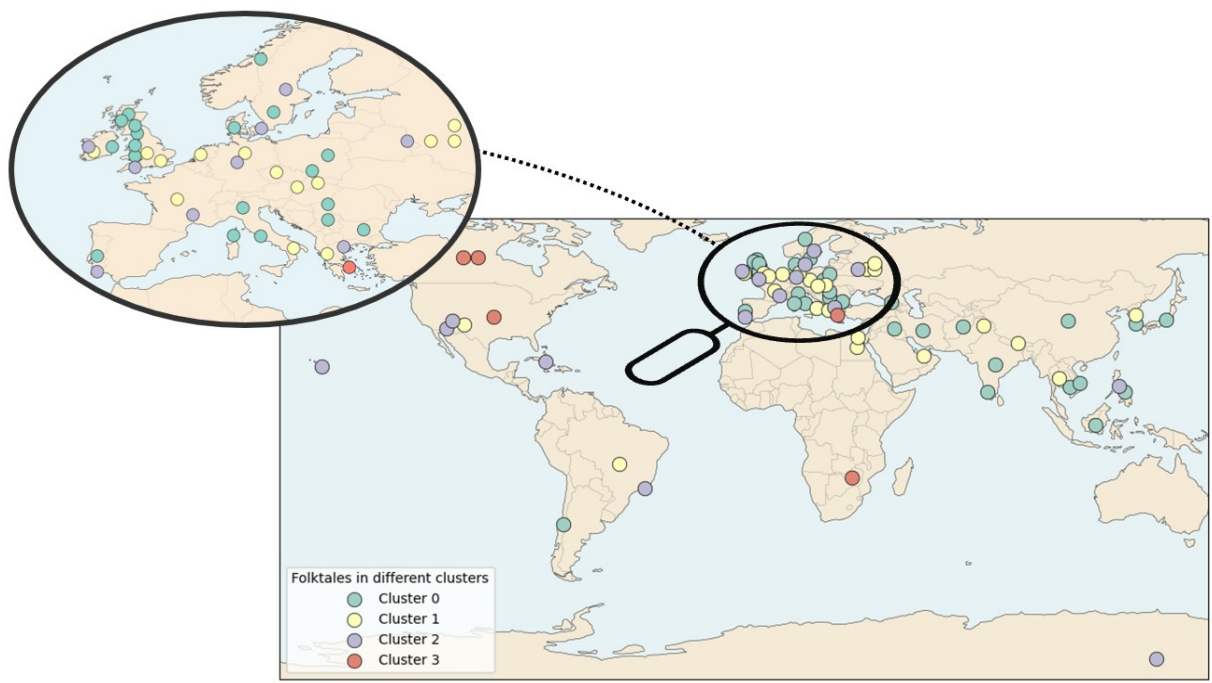

**Figure 9.** The geographical distribution of clusters based on broadly defined motifs

## 6 Discussion

While earlier folkloristic research relies on manual motif catalogues and tale type classification, this study demonstrates how combining traditional folkloristic approaches with computational methods can be used to advance research in folkloristics. We demonstrated that LLMs recognize diverse recurrent motifs in tales and identify deviations and subversions of individual narratives. These findings can be further used for a more nuanced analysis of different folktales and their variants, which we illustrated through the *Cinderella* case study. However, all subsequent computational analyses are based on the motifs we based our study on, which appears to be problematic. Our study demonstrated that existing motif categorizations do not adequately capture the full scope of folktale variants divided into different types, which limits the validity of our results.

Although we refined some of the original narrowly defined motifs and introduced new ones, our analysis still did not capture all the specific features of the narrative variants, but only those encompassed by the prescribed motifs. This limitation was intentional, as our aim was to test the validity of the existing tale type indexation; therefore, in defining motifs, we adhered to Thompson's motif index. Even when we generalized motifs, we drew on Thompson's superordinate categories, so as to remain consistent with the established indexing framework. For example, we altered the motif of helpful birds into helpful domestic animals and helpful wild animals, yet several other similar adjustments could have been made to address subtle differences among *Cinderella* variants. We found that LLMs were effective in identifying such deviations of individual narratives, indicating that they could be used successfully to refine the initial set of motifs for further analyses. This approach suggests alternative possibilities for identifying motifs in folktales.

Similarly, we tried to address the limitations of narrowly defined motifs by introducing broader supermotifs that encompass the ideas expressed by more specific motifs. While these generalizations were based on LLM suggestions, they were grounded in traditional motif categorizations. For example, we replaced the motif of a stepdaughter heroine with a victorious youngest child, which was the supermotif directly superordinate to the previous motif. However, this substitution again revealed the shortcomings of existing motif categorizations as the general motif failed to capture all types of child heroes but focused exclusively on youngest children who were victorious. As a result, narratives featuring a child who was neither the youngest nor victorious were excluded.

To achieve valid results with the help of computational methods, we need to define motifs in extreme detail, since LLMs interpret words quite literally. For example, we substituted the narrowly defined motif of clothes produced by magic with the broader motif of magic clothes in the general set. However, the meanings of the two motifs differ, and consequently, the LLM detected different distributions of the motifs across narratives. Similarly, the motif of flight required further clarification as hurried escape, since the term alone could carry multiple meanings.

Although the initial motifs that provided the basis for our analysis were not ideal, they nonetheless allowed us to test whether narratives can be divided into types on the basis of motifs using computational methods. Furthermore, we demonstrated how previously unanalysed and mostly unclassified Slovenian *Cinderella* variants in the Slovene language align with the non-Slovene clusters of *Cinderella* tales. In other words, our results show that motif-based clustering can be used to identify folktale types in new examples of narratives.

# 7 Conclusions

In this study, we propose a novel methodology for the automatic identification of tale types and motifs using large language models (LLMs). We demonstrate that models such as GPT-4.5-Preview can effectively detect recurring narrative motifs in tales. As each tale type is defined by a characteristic set of motifs, we can determine whether a given tale fits a specific type by identifying the motifs present in the text. This automatic approach enables a large-scale analysis of tale corpora and reduces the need for time-consuming manual annotation.

Additionally, our results show that LLMs are capable of capturing not only the presence of motifs but also their variations and deviations from established motif patterns. The LLM-based analysis of the motifs in *Cinderella* variants revealed limitations in the current motif typology, which is often too specific and fails to reflect the full range of narrative variation. This finding could potentially support the development of an automated data-driven folktale typology that is better aligned with motif patterns across diverse folktales and fairy tales.

We further demonstrate that the presence or absence of motifs, as identified by an LLM, provides a foundation for determining how similar examples of narratives are. Using embedding-based methodology we can cluster groups of similar tales, visualize their relationships, and enable a further interpretative analysis of each group of tales. The development of a novel methodology for automatically detecting tale types and motifs offers a valuable contribution to the field of computational folkloristics and opens new possibilities for large-scale narrative analyses.

While our approach shows that LLMs can successfully identify motifs in texts and group folktale variants into types, our results also demonstrate that traditional motif categorizations are inadequate. Further research is needed to address these limitations so that motifs used as the basis for computational methods can lead to more valid results. This study provides a foundation for future work focused on developing more systematic and computationally grounded frameworks for defining folktale types and their distinctive motifs.